\pdfoutput=1
\documentclass[format=sigconf,twocolumn,review=false,natbib=false]{article}
\usepackage[utf8]{inputenc}
\usepackage{soul}
\usepackage{graphicx}
\usepackage{booktabs}
\usepackage{multirow}
\usepackage{fullpage}
\usepackage{jlcode}
\usepackage[font=footnotesize,labelfont=bf]{caption}
\usepackage{subcaption}
\usepackage{booktabs}
\usepackage{framed}
\usepackage{blindtext}
\usepackage{url}
\title{Unsupervised Construction of\\ 
Knowledge Graphs From Text and Code}
\author{
  Kun Cao\\
  \texttt{Georgia Tech Research Institute}
  \and
  James Fairbanks\\
  \texttt{Georgia Tech Research Institute}
}
\renewenvironment{abstract}
 {\quotation\small\noindent\rule{\linewidth}{.5pt}\par\smallskip
  {\centering\bfseries\abstractname\par}\medskip}
 {\par\noindent\rule{\linewidth}{.5pt}\endquotation}
\date{}
\begin{document}

    \maketitle
    \begin{abstract}
        The scientific literature is a rich source of information for data mining with conceptual knowledge graphs; the open science movement has enriched this literature with complementary source code that implements scientific models.
        To exploit this new resource, we construct a knowledge graph using unsupervised learning methods to identify conceptual entities. We associate source code entities to these natural language concepts using word embedding and clustering techniques.

        Practical naming conventions for methods and functions tend to reflect the concept(s) they implement. We take advantage of this specificity by presenting a novel process for joint clustering text concepts that combines word-embeddings, nonlinear dimensionality reduction, and clustering techniques
        to assist in understanding, organizing, and comparing software in the open science ecosystem. With our pipeline, we aim to assist scientists in building on existing models in their discipline when making novel models for new phenomena. By combining source code and conceptual information, our knowledge graph enhances corpus-wide understanding of scientific literature.
    \end{abstract}

\section{Introduction} 
The corpus of scientific literature is growing exponentially, leaving individual researchers struggling to keep up. Natural language processing (NLP) techniques can help us understand this literature. However, as science becomes more dependent on large scale software, modeling and simulation, and data analysis scripts, the knowledge contained in publications shifts from the text of the papers to the software artifacts used to produce them. With the rise of open science and the drive to share open source code that implements scientific models, we are able to conduct analysis on scientific source code for the first time.

Source code definitions are typically semantic abbreviations of the scientific concepts they implement. We demonstrate that this constrained vocabulary is sufficient enough to create mappings to concepts extracted from natural language by using vectorized distances of word-embeddings~\cite{word2Vec}. Subsequently, we build knowledge representations that connect the conceptual relationships from scientific texts, with the procedural information embodied in open source scientific code. This paper proposes a knowledge graph framework for that knowledge representation, as well as a methodology for constructing said graph using both rule-based and unsupervised-learning techniques. Our methodology demonstrates an automated process of concept extraction using an open source textbook on epidemiological modeling and provides semantic meaning to code. These knowledge graphs can be used in future research to understand, organize, and augment scientific models.

\paragraph{Motivation}
The core motivation of this project is to support
SemanticModels.jl~\cite{SemanticModels}, which is a system that allows
scientists with limited scientific computing backgrounds to modify existing
implementations that are similar to their model. The knowledge graph of
reference text and code provides a method of searching for other software models
that are semantically similar. Knowledge graphs are generated for each model and
stored within a code base where similarity is determined through comparison of
conceptual nodes. This information, along with other data provided by dynamic and
static analysis, gives SemanticModels.jl the capability to detect similar models
and perform model transformations.

\paragraph{Related Work}
Related work includes the generation of coding comments with Deep Learning, which makes the assumption that the transition process between source code and comments is similar to the translation process between different natural languages~\cite{deepCC}. The Abstract Syntax Tree (AST) is used to model extracted concepts from source code to enable translation to natural language through traversal.
\emph{NSEEN: Neural Semantic Embedding for Entity Normalization} tackles a similar problem of constructing a knowledge graph from extracted text from various domains. They introduce a process called \emph{entity normalization}, which consists of mapping entity mentions from reference texts to another set of established entities from \emph{reference sets} through the use of Siamese recurrent neural networks~\cite{nseen}.

Text can also be converted to knowledge graph entities through the use of Long Short-Term Memory networks ~\cite{lstm,mapTextKG, embeddingModelsEntities}. Supervised learning through the use of Random Walk and a LSTM recurrent network is used to create skipgram entities that are included in a knowledge base. Input text is then assigned to these entities based on semantic similarity. This requires paired samples of text and knowledge graph entities to train the models. Leveraging unsupervised techniques, our model offers a solution that defines entities without a handcrafted ontology.

Developing domain specific software is a costly process; with a large code base it can be hard to understand how the pieces fit together and how those pieces of software relate to concepts in the application domain of interest. We take scientific software typically used in modeling applications as an interesting case because both the code and text are relatively sophisticated and contributing to the field requires deep knowledge of both software and science concepts. \textit{A Survey of Machine Learning for Big Code and Naturalness}~\cite{MLforBigCode} demonstrates application of bringing semantic meaning to code by utilizing the naturalness hypothesis, which argues: (1) software is a form of human communication; (2) software corpora have similar statistical properties to natural language corpora; and (3) these properties can be exploited to build better software engineering tools.

Our approach leverages code naming conventions for matching software and domain application conceptual entities into a unified knowledge graph without paired training examples. Like the semantic web, our model serves to create concept similarity relationships that involve the content of the resources rather than the bibliometric structure of the documents. Furthermore, our model offers a solution to the multidimensionality of the ontologies within scientific domains and tackles the inherent complexity of Big Code~\cite{semanticWeb}~\cite{ontologyBasedNLP}~\cite{MLforBigCode}. 

\vspace{5mm}
\section{Methodology}
\paragraph{Textual Explanations of Modeling Concepts}
Online or interactive textbooks are a novel creation of the open science movement. These textbooks are created when an author decides to combine text, data, figures, and code into an interactive textbook. Such books are published under a copy-left license on collaborative platforms such as \url{github.com}. 
This medium allows for simple text extraction as compared to alternate formats such as PDF and HTML. Variables, equations, and bibliographic references can be extracted from markdown documents with regular expressions that make text cleaning relatively easy.  For our model, we leverage the expository text and scientific or mathematical variables that are referenced in the markdown files of a textbook as shown in \textbf{Figure 1}. We decompose sentences from the reference text into \texttt{<subject, verb, object>} triples to form an RDF\footnote{resource description framework} knowledge graph. Edges flow from subjects to objects and are labeled with verbs. A sample of the resulting knowledge graph is shown in \textbf{Figure 2}.

%While equations from the markdown file could provide valuable information, we were unable to make use of them in our current pipeline.

\begin{figure}
\begin{center}
%\begin{framed}
\begin{subfigure}[t]{0.5\textwidth}
\includegraphics[width=\textwidth]{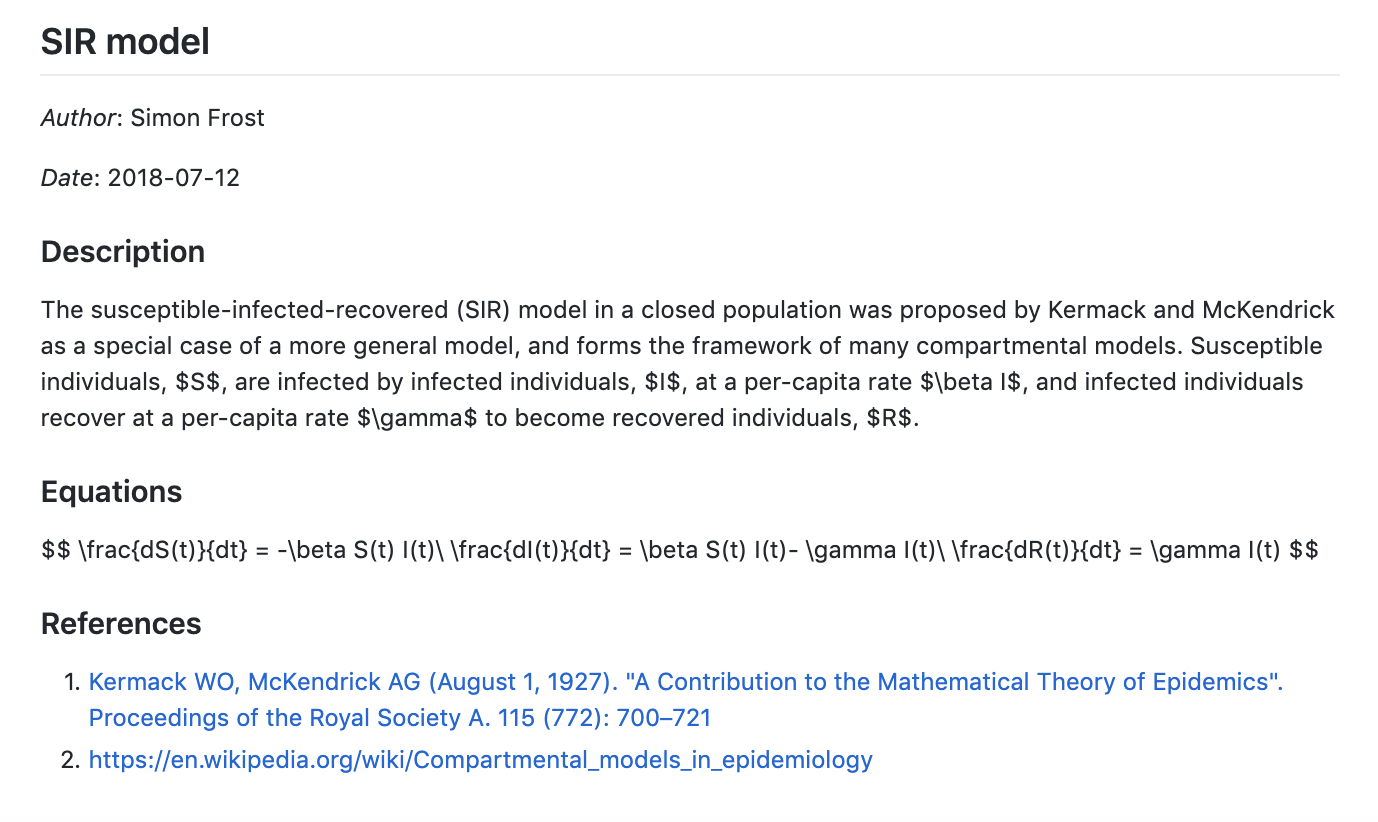}
\caption{}
\label{fig:textsnippet}
\center
\end{subfigure}
%\end{framed}
%\begin{framed}
\begin{subfigure}[t]{0.5\textwidth}

\includegraphics[width=\textwidth]{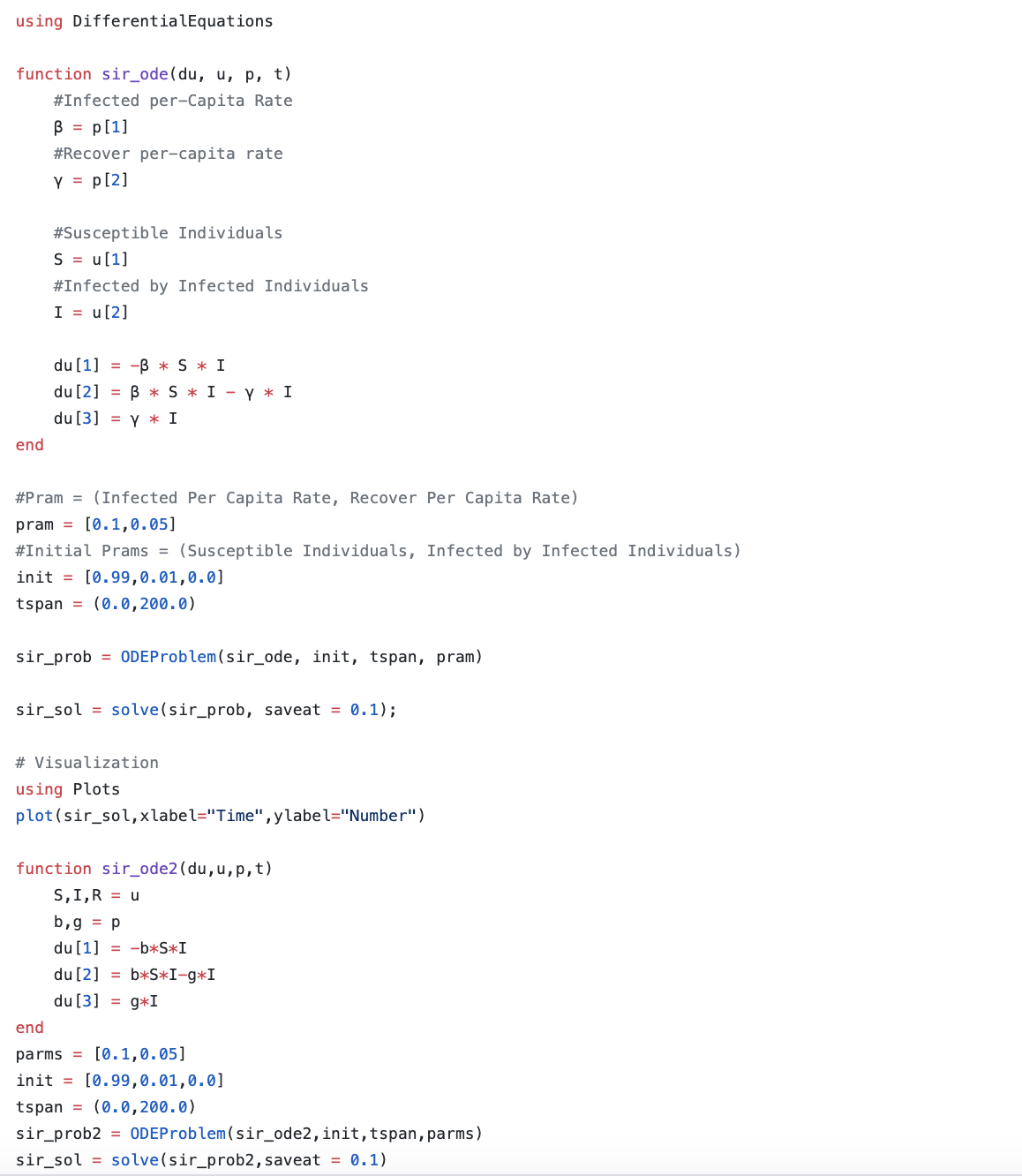}
\caption{}
\label{fig:sourcecode}
\center
\end{subfigure}
%\end{framed}
\caption{Modern online textbooks contain markdown files with expository text and Jupyter notebooks with code and figures. These input formats are designed for interactive instruction. Our work constructs scientific knowledge graphs for augmenting scientific reasoning from these data sources. \ref{fig:textsnippet}) Example of online \emph{Epirecipes Cookbook}~\cite{epicook} textbook markdown file with \ref{fig:sourcecode}) corresponding source code file. }
\label{fig:modeltypes}
\end{center}
\end{figure}

\paragraph{Code Implementations of Models}
%\begin{figure}[htbp]
%    \centering
%    \lstinputlisting{code.jl}
%    \caption{Models from the epirecipes cookbook look like %this...}
%    \label{fig:code_input}
%\end{figure}

When online textbooks are written to explain a scientific, engineering, or mathematical domain, examples are often given in the form of source code. These source code files are designed for pedagogical purposes and as such are well structured. These source code examples provide a high-quality, high-fidelity corpus for building knowledge graphs of the underlying domain. This allows for unambiguous association between concepts in the reference text and variable and function names extracted from code signatures. For example, a function named \texttt{sir\_ode()} can be  associated to an SIR concept referenced in the text. Furthermore, the code files associated with scientific models, being pedagogical, do not contain complex software constructions such as mutually recursive functions or low level functions manipulating complex data structures.  

\begin{figure}[!htb]
    \centering
    \includegraphics[width=0.5\textwidth]{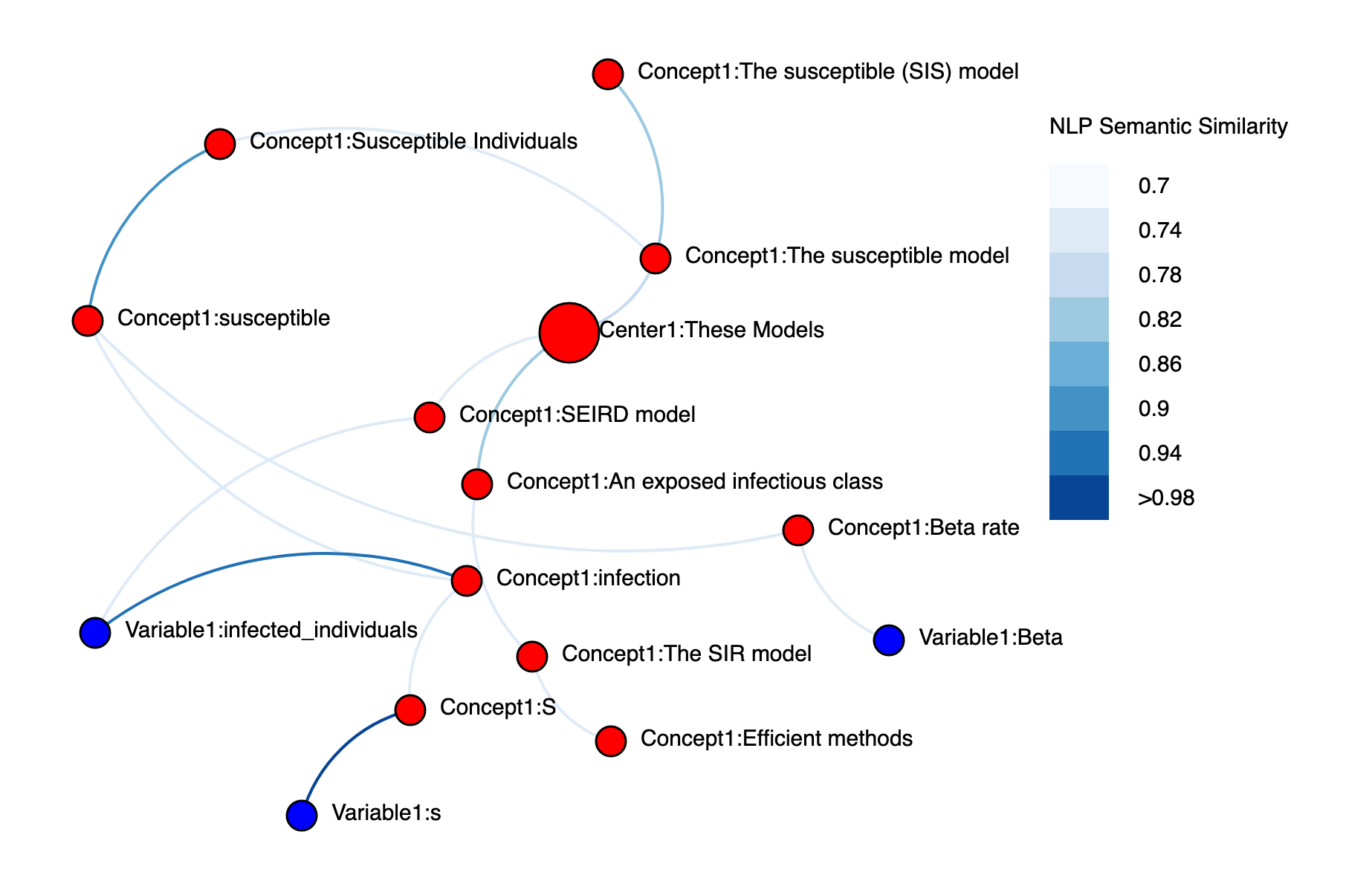}
    \caption{A small portion of the resulting knowledge graph. This portion of the knowledge graph shows the relationships between concepts in SIR modeling. The red vertices are concept nodes with the big red vertex representing a cluster center for concepts related to "These Models", and the blue nodes are source code variable nodes. In the bottom left of the figure, you can see that the {\color{blue}\texttt{infected\_individuals}} is related to the {\color{red}infection} concept which is related to the {\color{red}An exposed infectious class} concept. Additionally, the {\color{blue}\texttt{Beta}} variable is related to the {\color{red}Beta rate} concept which is related to the {\color{red}susceptible} concept.  }
    \label{fig:code_input}
\end{figure}

\paragraph{Pre-Processing}\textit{Epirecipes Cookbook}~\cite{epicook} is our primary source of data to build the knowledge graph since it provides us with a set of epidemiological models implemented in Julia and contains descriptive text about the models. Equations, references, and non-alphanumeric characters, with the exception of punctuation, are stripped from the text; remaining variables are then capitalized. Subsequently, subject, verb, and objects within sentences form source nodes, edges, and target nodes respectively for our knowledge graph through the use of the spaCy's small natural language processing model~\cite{spacy2}.

Functions and variables are then extracted from the Julia implementation corresponding to the models. Greek letter representations are translated to their respective Greek names. In future work, NLP models will be trained on large corpora of scientific texts. \footnote{\url{https://www.ncbi.nlm.nih.gov/pubmed/30217670}}

\vspace{5mm}
\section{Experimentation}
Our knowledge graph construction is based on creating clusters using word-embedding representations of the subject and object words from the cleaned text and then associate these variables and functions to representative elements from the object clusters. Rather than defining entities manually, we extract the entities from existing literature and source code. Therefore, we use the density-based spatial clustering of applications with noise (DBSCAN) to determine the number of clusters/entities for our model~\cite{dbscanSource}. 

Applying DBSCAN directly to the word-embeddings yields clusters that are too restrictive. The clusters contain text with only superficial variations in spelling and capitalization. For example in \textbf{Figure 3a}, phrases like ``The Model'', ``The Model of'', and ``The model in a closed population'' should share a semantic relationship but are not connected by the clustering algorithm. Tuning the DBSCAN epsilon parameter resulted in a reduction of the number of phrases that are clustered as noise, but did not resolve the problem of combining \emph{semantically similar} phrases.

\begin{figure}
\begin{center}
\begin{subfigure}[b]{.49\textwidth}
\includegraphics[width=\textwidth]{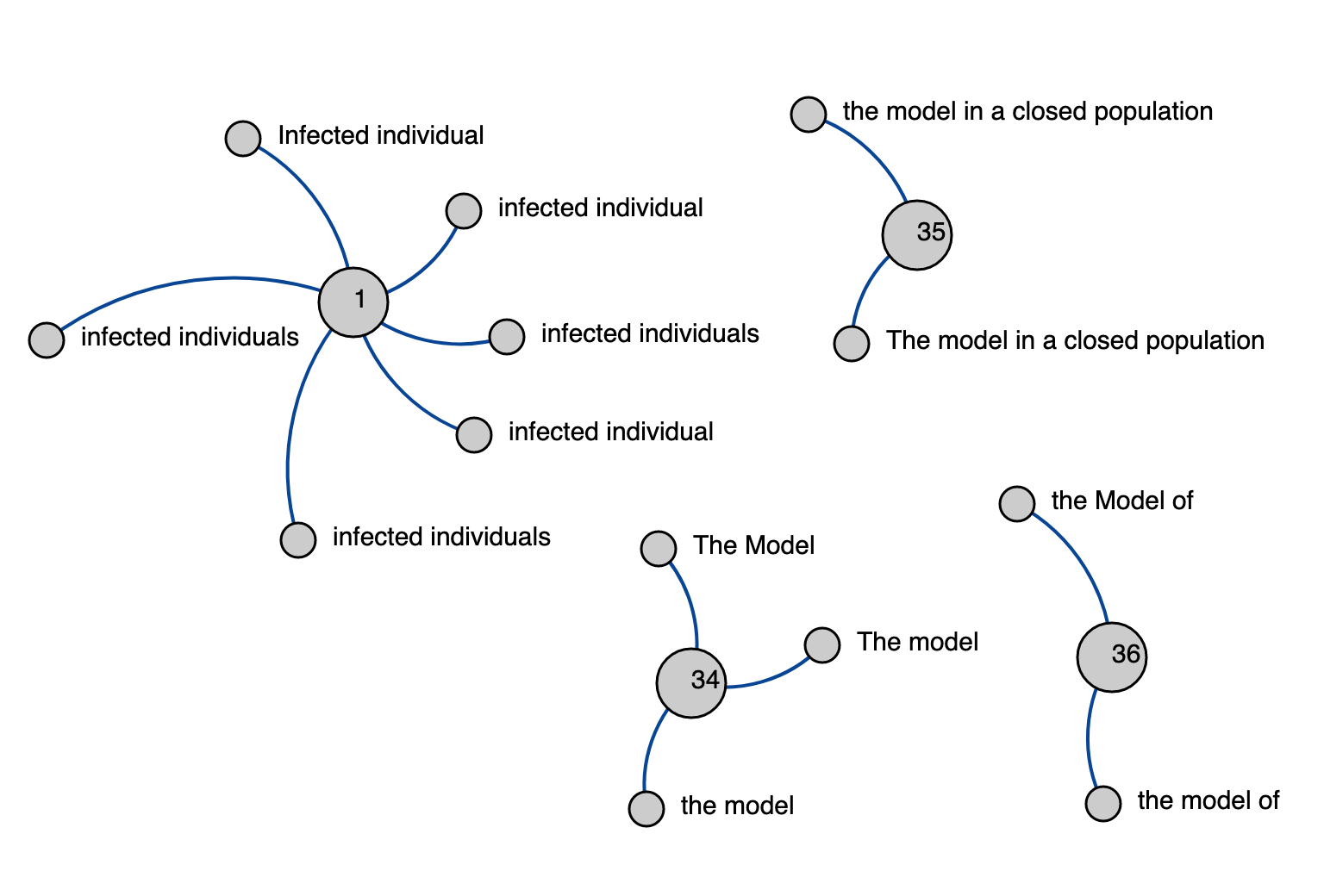}
\caption{}
\label{fig:clustersA}
\center
\end{subfigure}

\begin{subfigure}[b]{.49\textwidth}
\includegraphics[width=\textwidth]{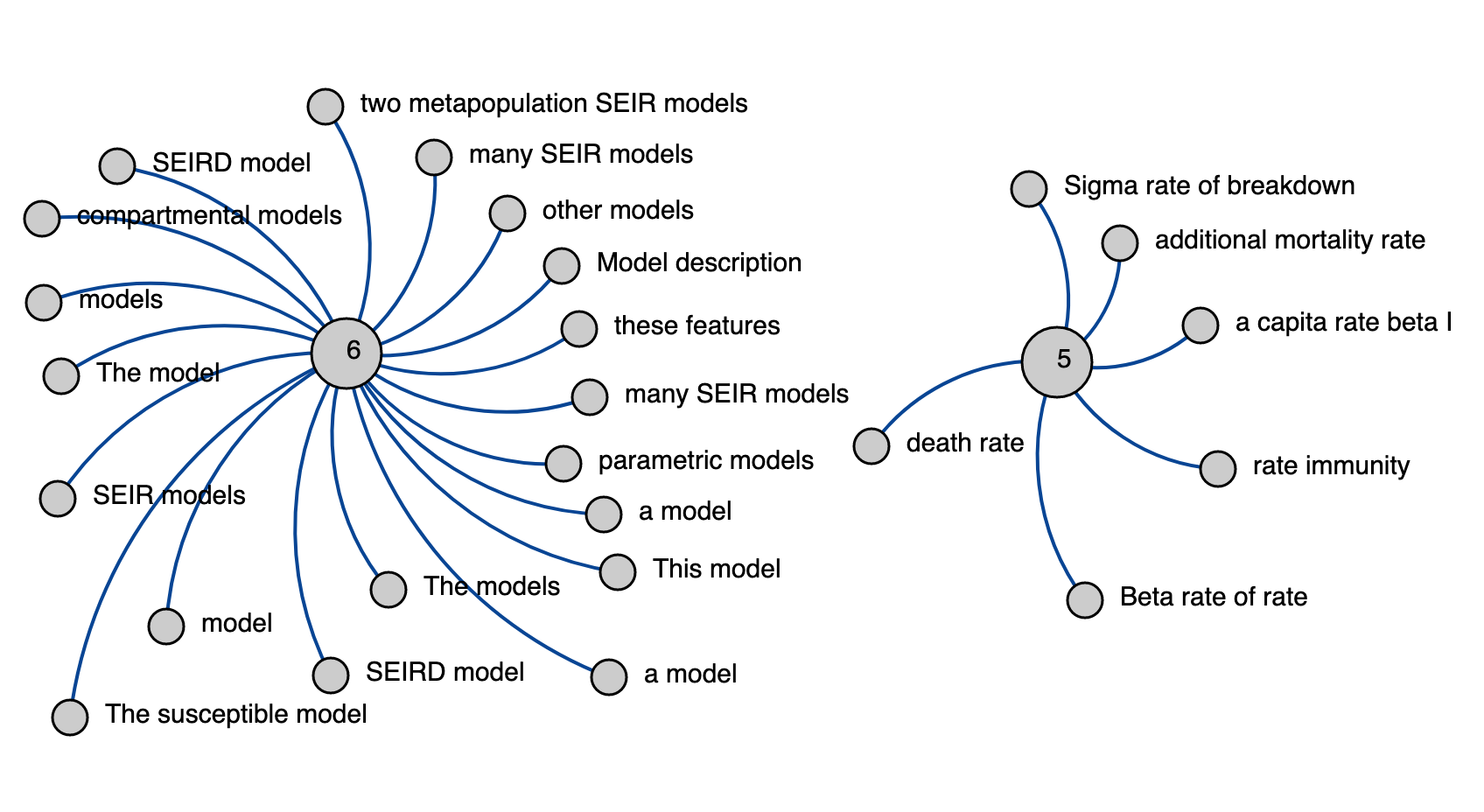}
\caption{}
\label{fig:clustersB}
\center
\end{subfigure}
\caption{A comparison of clusters with and without dimensionality reduction. \textbf{3a} Cluster Assignments for high dimensional word-embeddings. The high dimensional space separates vectors well and each cluster contains only superficial variations on the same phrase. \textbf{3b} UMAP transformation embeddings resulting in semantically significant clusters. This is because the nonlinear embedding of the vectors provided by UMAP pulls phrases together in the lower dimensional space. These clusters are at a resolution too fine for the application of knowledge graph construction.}
\label{fig:modeltypes2}
\end{center}
\end{figure}

To tackle this problem, we apply a UMAP transformation to the word-embedding to reduce the dimensionality of the input space for clustering~\cite{umap}. As a result with a DBSCAN epsilon value of 0.30, the clusters in the UMAP embedded space compose phrases that are diverse in lexical level, but similar as concepts. In \textbf{Figure 3b}, various types of models are assigned to cluster 6, whereas types of rates are assigned to cluster 5. 

However, because the function and variable names are specific, it is difficult to associate our extracted code signatures with these high-level concepts. Therefore, we use DBSCAN with UMAP transformation only on the subject nodes and exclude the noise to create our hierarchical entities. Then, we use DBSCAN without a UMAP transformation on the objects to combine syntactically similar nodes. This captures an intuitive sense that there are more objects than subjects in the corpus. Finally, we connect objects (with noise) to subject entities based on their original \texttt{<subject, verb, object>} association. Subsequently, extracted variables and functions are compared to the object nodes and are connected when the similarity exceeds a fixed threshold discussed in Section 4. 

The resulting knowledge graph created from the \emph{Epirecipes Cookbook} yielded 115 object nodes and 93 subject nodes. Depending on the threshold, the number of edges ranged from 4,000 to 13,000. In order to remove extraneous concepts, subject components with a node size of 5 or lower were removed from the knowledge graph.

\vspace{5mm}
\section{Results and Discussion}

The construction of the knowledge graph is dependent on factors such as the threshold value and the input resource. In this section, we introduce methodologies of determining the threshold value through precision versus recall and measures of conductance when including an additional scientific corpus.

Internal performance measures, such as Silhouette index~\cite{silhouettes} and other intra-cluster similarity assessments would not provide accurate evaluation for our model given examples like \textbf{Figure 3a}, where the Silhouette index would be high. Other evaluation metrics, such as measuring the purity of the cluster, would also not apply since we aim to compare and group shared functionality within a single scientific class-epidemiology.

In order to assess the threshold value for our variable assignment, we crafted a set of ground truth labels that were hand-labeled by a group of peers. These labels were created from a list of object nodes that reflect the variable/function nodes that they should be connected with. Evaluation was conducted with respect to these labeled sets in terms of precision versus recall at various thresholds (see \textbf{Figure 4}) where $Precision = {tp}/{(tp+fp)}$, and 
$Recall = {tp}/{(tp+fn)}$.

\begin{figure}[!htb]
    \centering
    \includegraphics[width=0.49\textwidth]{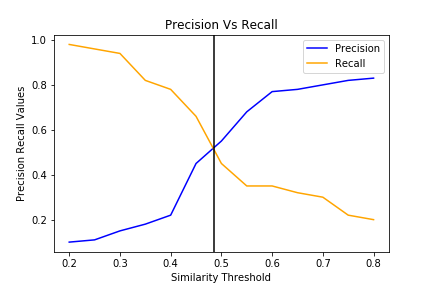}
    \caption{Precision vs Recall trade-off in this context. Knowledge graph construction often prefers high recall, low precision thresholds because false positives can be filtered out in the downstream learning or reasoning steps.} 
    \label{fig:Similarity_Threshold}
\end{figure}

In this application,
\textbf{true positives} are \texttt{<variable/function, object>} edges that exist in the knowledge graph and in the labeled set. \textbf{False positives} are \texttt{<variable/function, object>} edges that were added to the knowledge graph, but not in the labeled set. \textbf{False negatives} are \texttt{<variable/function, object>} edges that do not exist in the knowledge graph, but are in the labeled set.

As the similarity threshold increased, the number of edges between variable and object concepts are decreased. In order to determine the threshold value to use, we first discovered the threshold value where recall was equivalent to precision. For our model, we chose a threshold higher than the intersection point because extraneous edges can be filtered out later in a downstream processing task. We selected a threshold value of 0.7 to give a good balance between precision and recall for our knowledge graph applications.

\textbf{Figure 2} shows a snippet of the constructed knowledge graph; subject \emph{concept} nodes are connected to other object \emph{concept} nodes extracted from our \texttt{<subject, verb, object>} triplet from the reference text. Variable names as a result were connected to these object nodes through satisfaction of our similarity threshold parameter. The amount of variable matches we obtained in our graph was greatly dependent on the quality of the language triplets extracted from the reference text. For example, the variable \texttt{infected\_individuals} would not have been built into our knowledge graph had the triplet \texttt{<susceptible, contains, infection>} not existed. Our pipeline mimics human learning in the sense that the more concept associations that are present, the more likely it is able to draw connections and make semantic sense from source code.

\begin{figure}[!htbp]
  \centering
  \includegraphics[width=0.5\textwidth]{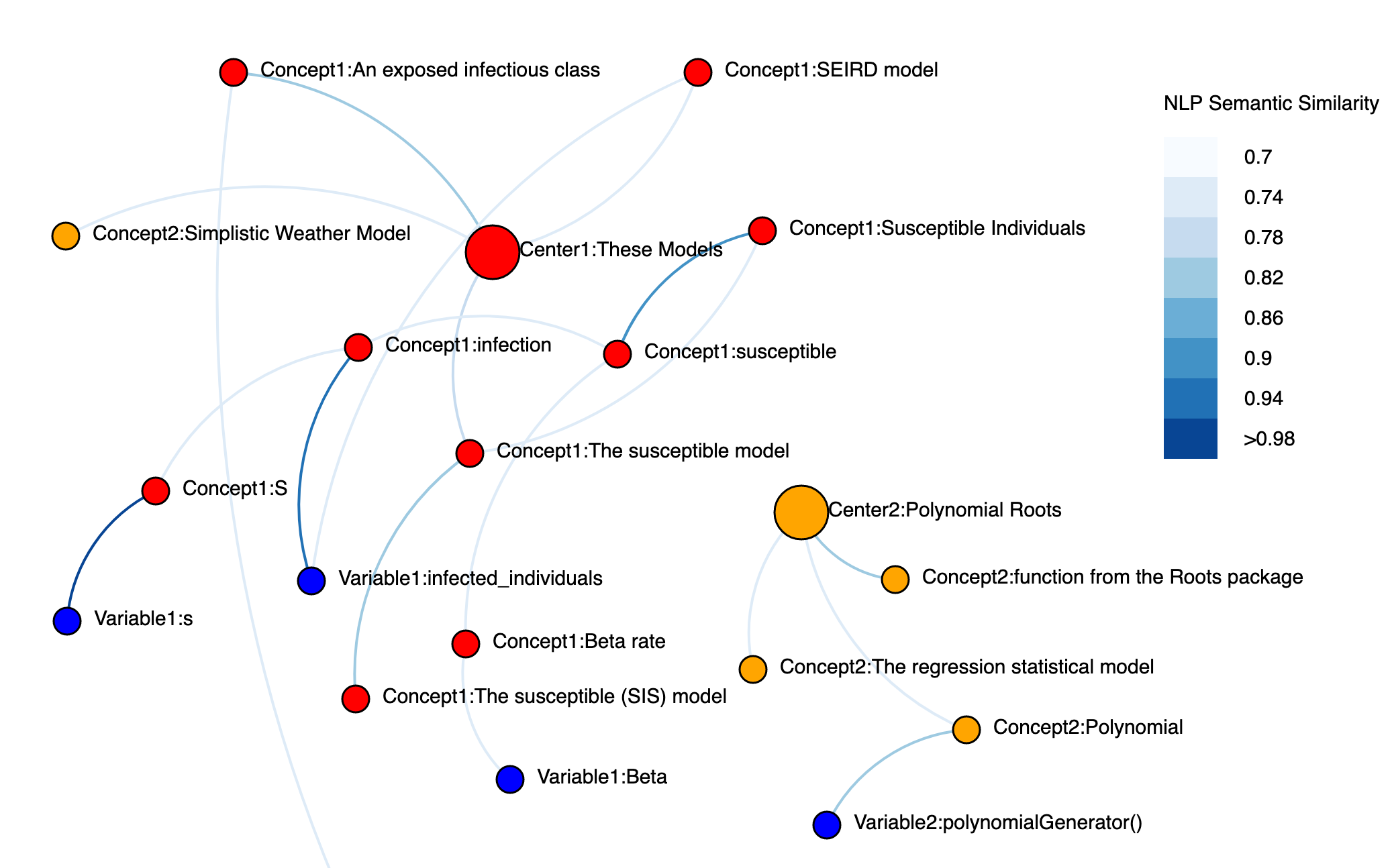}
  \caption{A portion of the knowledge graph extracted from two online textbooks: \emph{Epirecipes Cookbook}~\cite{epicook} and  \emph{Statistics with Julia: Fundamentals for Data Science, Machine Learning and Artificial Intelligence}~\cite{juliaStats}. Note that the subgraphs corresponding to each textbook (Epirecipes in red and Statistics in green) are mostly separated, but there are some inter-textbook connections namely \texttt{<\textcolor{orange}{Simplistic Weather Model Model}, \textcolor{red}{These Models}>}. These textbooks both talk about modeling physical phenomena with mathematics and so we should expect overlap.}
    \label{fig:Similarity_Threshold2}
\end{figure}

Furthermore, \textbf{Figure 5} demonstrates a knowledge graph that introduces a new corpus from the textbook,
\textit{Statistics with Julia: Fundamentals for Data Science, Machine Learning and Artificial Intelligence}~\cite{juliaStats}.
The new introduction produces an independent set demonstrating that the two textbooks were relatively disjointed in terms of similarity.
The construction of our graph allowed for quantitative assessments to be performed to evaluate similarity of these two resources.
Given $G=(V,E)$ to represent our knowledge graph where $V_a\subset V$ represents the set of vertices produced by a corpus $a$ and $V_b\subset V$ represents the set of vertices introduced by a new corpus, $b$. The sets $V_a, V_b$ form a partition of the vertex set $V$.
Let $N_a$ and $N_b$ denote the size of $V_a$ and $V_b$ respectively.
The fraction of the knowledge graph extracted from corpus $a$ is calculated as ${N_a}/{(N_a+N_b)}$.
The conductance of any vertex partition measures the separation between these vertices in terms of path connectivity. When combining document corpora into a single knowledge graph, the conductance reflects how many connections between the domains of each corpus were introduced by the knowledge graph constructions algorithm. The conductance of a cut $S, V\setminus S$ in graph $G$ is
\[
          \phi{(S)} = \frac{\sum_{i \in S, j\in \bar{S}} a_{ij}}
          {a(S)}
\]
% \[
%   \min_{S\subseteq V; 0 \leq a(S) \leq \frac{a(V)}{2}}
%           \frac{\sum_{i \in S, j\in \bar{S}} a_{ij}}{a(S)}
% \]
where $a_{ij}$ are the entries of the adjacency matrix for $G$ such that:
\[
  a(S) = \sum_{i \in S} \sum_{j \in V} a_{ij}
\]
and $\bar{S} = V\setminus S$. 

When combining two corpora, for example two textbooks on the same or different topics, one can define $S$ as the set of concepts and variables extracted from one corpus and compute the conductance $\phi{(S)}$.
\textbf{Table 1} shows the conductance $\phi{(S)}$ between the epidemiology concepts and the statistics concepts in a knowledge graph built by extracting from two textbooks.
The results show that an increase in the similarity threshold decreases the number of variable-object relationships across disciplines. The decline of the number of edges lowers the conductance $\phi(S)$.

\begin{table}[htbp]
\centering
\begin{tabular}{@{}rr@{}}
\toprule
Threshold         & Conductance \\ \midrule
0.20              & 0.0285     \\
0.25              & 0.0152     \\
\textbf{0.30}              & \textbf{0.0114}     \\
\textbf{0.35}              & \textbf{0.0090}     \\
0.40              & 0.0089     \\
0.45              & 0.0087     \\
0.50              & 0.0086     \\
0.55              & 0.0085     \\
0.60              & 0.0085     \\
0.65              & 0.0085     \\
0.70              & 0.0084     \\
0.75              & 0.0084     \\ 
0.80              & 0.0084     \\\midrule
\end{tabular}
\caption{
  The transition point between the threshold value 0.30 and 0.35 (indicated in bold) corresponds to a large gap in conductance (0.0114 to 0.0090) and illustrates a method for choosing a threshold. A lower threshold reflects a higher conductance, that is, more interdisciplinary edges in the network. While these edges are useful for discovering relationships between disparate scientific disciplines, too many of them indicates an imprecise understanding of the concepts within a single discipline. 
  }
\end{table}

\vspace{5mm}
\section{Conclusion and Future Work}

Semantic modeling aims to extract scientific knowledge that reside in scientific code. With the abundance of open source code and reference texts, our model provides knowledge to reduce the complexity of large code bases to assist scientists and developers in gaining an overview understanding of the source code. Our framework based on unsupervised learning and word-embeddings does not require a large volume of ontological examples and enables the extension of models with new parameters and components. 

\paragraph{Future Work}
The performance of our model relies on the quality of  function and variable names. Storage and placeholder variables extracted from source code diminish the accuracy of our model due to terse initialization names. For our future work, we aim to handle these semantically insignificant variables through methods of classification.
Furthermore, because the word-embeddings are trained on the English language corpus rather than coding syntax, we run the risk of lexical entities extracted from reference text skewing our results. But because of the high-quality of naming schemes presented in the \emph{Epirecipes Cookbook}, the association between concept and code is less ambiguous.

\paragraph{Applications}
Software engineering and scientific software development are high turnover fields where people rotate onto projects and must come up to speed quickly. When new engineers join a project or new scientists add a new software method to their repertoire, they must first gain understanding of what is already implemented. An application of our model can greatly assist in this transitional period by providing insight into existing code-bases. Knowledge graphs constructed from software and documents constructed using our methods can support semantic software engineering applications.

Additionally, artificial intelligence systems designed to augment the performance of scientists will need a deep understanding of the domain science of interest. This understanding can be constructed from knowledge graphs built by reading textbooks and code in the scientific domain. By supporting the construction of domain specific knowledge graphs, these methods can contribute to the next generation of methods for applying machine learning to aid in scientific discovery.

\vspace{5mm}
\section{Acknowledgments}
The authors thank the authors of \emph{Epirecipes Cookbook} and \emph{Statistics with Julia:
Fundamentals for Data Science, Machine
Learning and Artificial Intelligence}; without these sources, this work would not exist. We also thank Christine Herlihy, Kevin Kelly, and Clayton Morrison for their advice on this manuscript. This material is based upon work supported by the Defense Advanced Research Projects Agency (DARPA) under Agreement No. HR00111990008.

\newpage
\bibliographystyle{plain} % Plain referencing style
\bibliography{ref} % Use the example bibliography

\end{document}